\documentclass[pdflatex,sn-mathphys-num]{sn-jnl}% Math and Physical Sciences Numbered Reference Style
\usepackage{graphicx}%
\usepackage{multirow}%
\usepackage{amsmath,amssymb,amsfonts}%
\usepackage{amsthm}%
\usepackage[title]{appendix}%
\usepackage{xcolor}%
\usepackage{textcomp}%
\usepackage{manyfoot}%
\usepackage{booktabs}%
\usepackage{algorithm}%
\usepackage{placeins}
\usepackage{algorithmicx}%
\usepackage{algpseudocode}%
\usepackage{listings}%
\usepackage[most]{tcolorbox}

\usepackage[inline]{enumitem}
\usepackage{xcolor}
\usepackage{placeins} 
\theoremstyle{thmstyleone}%
\theoremstyle{thmstyletwo}%

\theoremstyle{thmstylethree}%

\tcbuselibrary{breakable}

\definecolor{angercolor}{HTML}{D55E00}
\definecolor{disgustcolor}{HTML}{E69F00}
\definecolor{fearcolor}{HTML}{7CAE59}
\definecolor{happinesscolor}{HTML}{E9C46A}
\definecolor{sadnesscolor}{HTML}{6C8EBF}
\definecolor{surprisecolor}{HTML}{7B6FB6}

\newcommand{\ANGER}{\textcolor{angercolor}{angry}}
\newcommand{\DISGUST}{\textcolor{disgustcolor}{disgusted}}
\newcommand{\FEAR}{\textcolor{fearcolor}{afraid}}
\newcommand{\HAPPINESS}{\textcolor{happinesscolor}{happy}}
\newcommand{\SADNESS}{\textcolor{sadnesscolor}{sad}}
\newcommand{\SURPRISE}{\textcolor{surprisecolor}{surprised}}

\definecolor{cotcol}{HTML}{FFF9DB}
\definecolor{vanillacol}{HTML}{E6F0FF}

\begin{document}

%\title{Emotion as Context: A Multi-Domain Evaluation of Large Language Models}

\title{Do Emotions in Prompts Matter? Effects of Emotional Framing on Large Language Models}

\author[1]{\fnm{Minda} \sur{Zhao}}\email{mindazhao@hsph.harvard.edu}
\equalcont{These authors contributed equally to this work.}

\author[1]{\fnm{Yutong} \sur{Yang}}\email{yutongyang@hsph.harvard.edu}
\equalcont{These authors contributed equally to this work.}

\author[1]{\fnm{Chufei} \sur{Peng}}\email{chufeipeng@g.harvard.edu}
\equalcont{These authors contributed equally to this work.}

\author[1]{\fnm{Rachel} \sur{Gonsalves}}\email{rgonsalves@mba2027.hbs.edu}
\equalcont{These authors contributed equally to this work.}

\author[1]{\fnm{Weiyue} \sur{Li}}\email{weiyueli@fas.harvard.edu}
\equalcont{These authors contributed equally to this work.}

\author[1]{\fnm{Ruyi} \sur{Yang}}\email{ruyi\_yang@gsd.harvard.edu}

\author[2]{\fnm{Zhixi} \sur{Liu}}\email{lliu2@brynmawr.edu}

% Corresponding authors:

\author*[1]{\fnm{Mengyu} \sur{Wang}}\email{mengyu\_wang@meei.harvard.edu}

\affil[1]{\orgname{Harvard University}}
\affil[2]{\orgname{Bryn Mawr College}}


%\abstract{%
%Real-world queries to large language models often carry emotional tone, yet how affective framing shapes model behaviour remains unclear. We manipulated \wl{change to present tense?} emotional content in prompts across six benchmark datasets. These datasets span mathematical reasoning, general reasoning, medical and commonsense question answering, reading comprehension and social inference. We evaluated Qwen3-14B, LLaMA-3.3-70B, DeepSeek-V3.2 and GPT-OSS \wl{can just say diverse open source models from different major providers?}. Prepending short emotional sentences to otherwise identical questions produced only small average accuracy shifts, ranging from zero to about minus 0.7 percentage points depending on task; medical question answering (MedQA-US) showed no mean change \wl{why do we want to mention medqa here?}. Effects varied by model and emotion rather than following a single pattern. On a held-out MedQA-US subset, we compared prompts prefixed with human written or model generated emotional sentences under the same model (Qwen3-14B). Accuracy under human injected emotion matched or slightly exceeded that under LLM injected emotion for most conditions. Emotional context thus acts as a mild, non systematic perturbation, and human authored affective framing can be used without systematically reducing performance. \wl{should also mention rl agent here?} \minda{Rewrite the whole section}
%}

\abstract{Emotional tone is pervasive in human communication, yet its influence on large language model (LLM) behaviour remains unclear. Here, we examine  how first-person emotional framing in user-side queries affect LLM performance across six benchmark domains, including mathematical reasoning, medical question answering, reading comprehension, commonsense reasoning and social inference. Across models and tasks, static emotional prefixes usually produce only small changes in accuracy, suggesting that affective phrasing is typically a mild perturbation rather than a reliable general-purpose intervention. This stability is not uniform: effects are more variable in socially grounded tasks, where emotional context more plausibly interacts with interpersonal reasoning. Additional analyses show that stronger emotional wording induces only modest extra change, and that human-written prefixes reproduce the same qualitative pattern as LLM-generated ones. We then introduce EmotionRL, an adaptive emotional prompting framework that selects emotional framing adaptively for each query. Although no single emotion is consistently beneficial, adaptive selection yields more reliable gains than fixed emotional prompting. Together, these findings show that emotional tone is neither a dominant driver of LLM performance nor irrelevant noise, but a weak and input-dependent signal that can be exploited through adaptive control.}



\maketitle

\section{Introduction}\label{sec1}

Large language models (LLMs) have become ubiquitous across domains including education, coding assistance, creative writing, and personal productivity \cite{ liang2025widespread,li2026llm, xu2024large}. As LLM-generated text increasingly appears in consumer communications, corporate messaging, and international correspondence \cite{liang2025widespread}, understanding how these systems process and respond to human communication has become critical. LLMs are trained on large-scale human language corpora and learn statistical patterns of human communication, yet they do not operate as neutral systems \cite{Wang2023}. This recognition has driven extensive research into LLM behaviors, including hallucination, fairness, toxicity, sycophancy, and consistency \cite{Gallegos2024}

Human communication is inherently emotional. Individuals convey affect through tone, framing, and lexical choice when expressing needs, negotiating disagreements, or seeking advice \cite{Barrett2017,Flusberg2024}. Consequently, real-world human--LLM interactions often carry an emotional tone that, even when unstated, shapes model outputs and influences how users interpret responses \cite{Riley2025}. Emotion is therefore central rather than peripheral to human--LLM interaction.

Emotion plays a causal role in human decision-making. Emotional states guide attentional focus, bias interpretation under ambiguity, and influence reasoning in uncertain environments \cite{forgas1992affect, Lerner2015}. These effects can alter behavior even when material incentives remain unchanged. Empirical work demonstrates that emotion functions as a causal mechanism linking context to behavior. For example, Heffner et al.\ \cite{Heffner2021} studied emotion prediction errors in social decision-making tasks involving punishment and forgiveness. They showed that discrepancies between expected and experienced emotion predicted subsequent decisions beyond reward-based prediction errors, isolating a pathway through which emotion updates future behavior and supports adaptive social interaction. Related work has shown that emotion-guided attention impacts perceptual decision-making \cite{Ngai2025}, emotional expressions influence visual search efficiency \cite{Olatunji2011}, and affective information integrates into experience, judgment, and decision-making \cite{Asutay2024}. Because LLMs are trained on emotionally mediated human text, affective tone in prompts may similarly modulate model behavior. However, the causal mechanisms through which emotional cues influence LLM performance remain underexplored.

Prior work has primarily studied emotion in LLMs through explicit prompt manipulations. In strategic settings, assigning emotional roles or varying task framing has been shown to shift model behavior in cooperation, bargaining, and related game scenarios \cite{Mozikov2024,Lore2024}. Related work has further shown that emotionally charged descriptions can alter norm-enforcement judgments, leading models to impose stronger punishment in altruistic punishment settings \cite{Wang2025Outraged}. Beyond strategic decision-making, several studies have reported that affective prompts can influence downstream task performance, including EmotionPrompt-style approaches designed to improve results across benchmarks \cite{Li2023EmotionStimuli,Wang2024NegativePrompt,Cheng2025}. Other lines of work have examined emotional expression in LLMs, responses to conflicting emotional cues, controlled affective text generation, and fine-grained emotion resources \cite{Ishikawa2025,UlHuda2024,Singh2020,Demszky2020}. Taken together, these studies show that LLMs are responsive to explicitly inserted emotional content. However, three important questions remain unresolved. First, it is still unclear whether emotionally framed prompts expressed in a first-person, user-voiced form produce systematic and reliable changes in model behavior across standard tasks. Second, prior work has not directly tested whether these effects depend on prompt authorship, particularly whether they hold for human-written emotional prefixes as well as LLM-generated ones. Third, existing studies have largely treated emotion as a fixed prompting condition rather than learning how emotional framing should be selected for a given input. These questions are particularly important given documented biases in LLMs \cite{Bai2025,Mahajan2025,Mao2022}, including cognitive biases in clinical settings \cite{Ke2024,Templin2025}, stigmatizing language in medical records \cite{Barcelona2024,Harrigian2023}, and fairness concerns \cite{Radaideh2025}.

In this work, we study how first-person emotional framing in user prompts affects LLM behavior. Rather than assigning emotions to the model or relying on coarse positive-negative distinctions, we prepend short first-person emotional prefixes that express the user’s affective state and examine six emotion categories: happiness, surprise, fear, sadness, disgust, and anger, guided by Plutchik’s basic emotions framework \cite{Plutchik1980} and Russell’s circumplex model of affect \cite{Russell1980}. We study these effects across reasoning and knowledge-based tasks, focusing on how task structure and complexity shape emotional modulation. Our contributions are threefold. First, we show that static emotional prompting is not a reliable general-purpose intervention: across benchmarks, fixed emotional prefixes typically induce only small average accuracy shifts, with effects that vary substantially across tasks, backbone models, and emotion categories rather than following a uniform pattern. Second, we demonstrate that this overall pattern is robust to targeted validation studies: increasing emotional intensity produces only modest additional change without creating a qualitatively different regime, and human-written prefixes reproduce the same qualitative conclusions as LLM-generated ones. Third, we introduce EmotionRL, an adaptive emotional prompting framework that formulates emotional prompting as an input-conditioned decision problem, and show that different questions benefit from different emotional framings. By selecting a more suitable emotion for each input, EmotionRL yields more reliable gains than either the no-emotion baseline or the average of fixed emotional prompts.

\section{Methodology and Experiment Setup}\label{sec11}

% \minda{I am not quite sure if we need this part.}This section describes the emotion taxonomy, datasets, models, prompting schemes, data preparation and evaluation protocol used in this study. The objective is to characterize how emotional context modulates the behaviour of large language models under controlled conditions and to enable exact replication of all experiments. Emotional context is treated as an experimental manipulation of the input prompt. Two complementary mechanisms are used. First, an external stimulus based manipulation adds emotionally valanced sentences to otherwise neutral prompts. An internal paraphrasing based manipulation rewrites prompts to express a target emotion while preserving task relevant semantics. Both mechanisms are grounded in a psychologically motivated emotion taxonomy and are applied consistently across models and benchmark tasks spanning reasoning, knowledge retrieval and domain specific question answering. A shared evaluation pipeline is used to compare behaviour across emotion conditions, tasks and model architectures.

\begin{figure}[H]
  \centering
  \includegraphics[width=\linewidth]{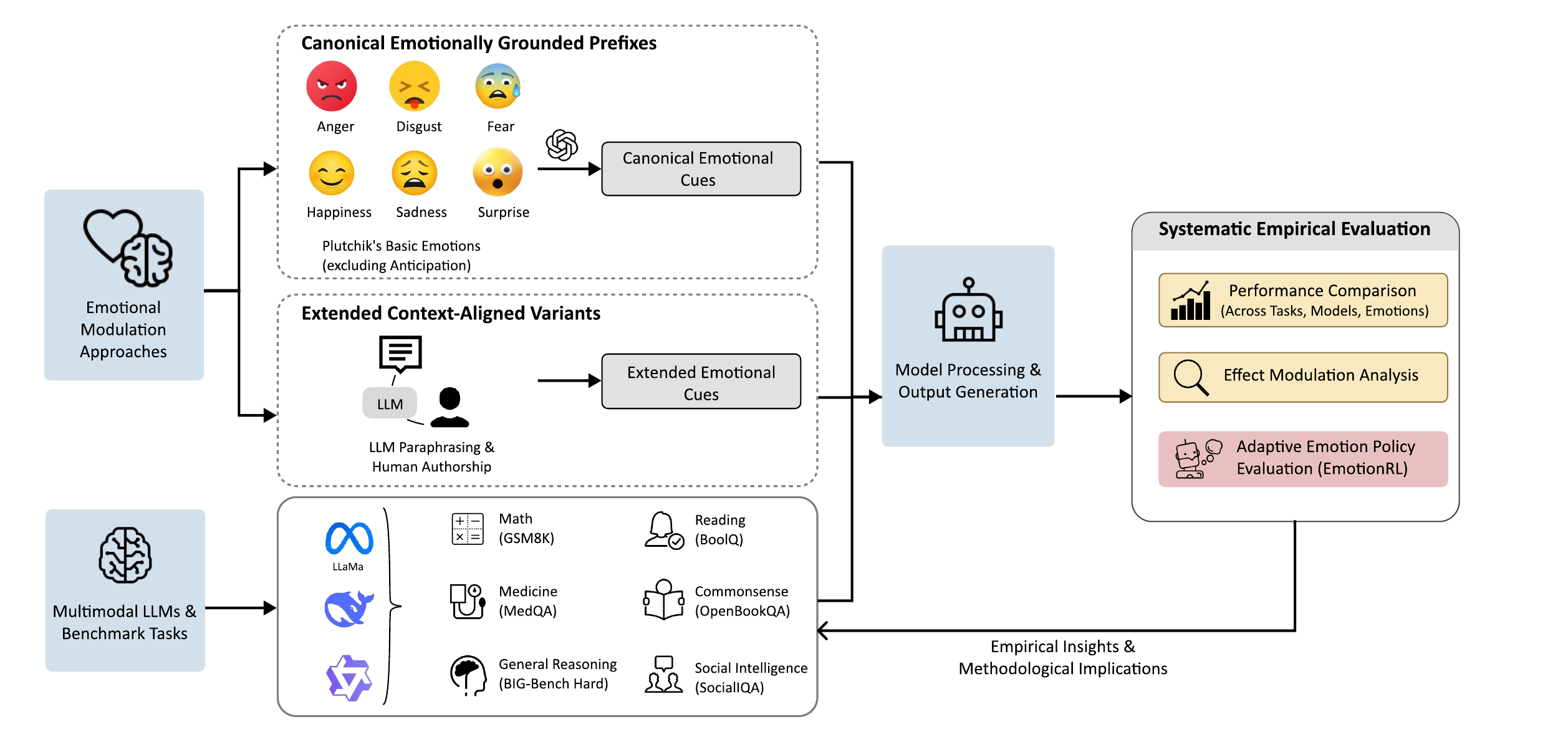}
  \caption{Pipeline overview. We generate emotion-conditioned inputs via canonical prepended cues and context-aligned variants, evaluate them with frozen LLMs across diverse benchmarks, and analyze performance sensitivity across models, tasks, and emotions. We also include adaptive per-instance emotion selection with EmotionRL.}
  \label{Fig emotion framework}
\end{figure}

% \subsection*{Study design}

%The study examined how emotional content in prompts alters the behaviour of large language models across a range of cognitive and knowledge based tasks. Emotional manipulations were applied to the textual input while the underlying tasks and reference answers remained constant. Three input conditions were defined for each item. The neutral condition presented the original question without added emotion. The external emotion condition prepended a short emotional sentence to the question. The internal emotion condition replaced the question with an emotion infused paraphrase that preserved all task critical details. Models were treated as experimental units exposed to matched sets of stimuli. All models answered the same items under fixed prompting and decoding settings for a given dataset. The use of shared prompts response formats and scoring rules across conditions was intended to reduce confounds and to support direct behavioural comparisons across models emotions and manipulation types.

\subsection{Design and implementation of affective manipulations}

\paragraph{Emotion taxonomy.}
Emotion conditions were operationalized using a discrete taxonomy informed by established theories of affective organization. We drew on Plutchik's basic emotions framework \cite{Plutchik1980} and Russell's circumplex model of affect \cite{Russell1980} to define six emotion categories: happiness, sadness, fear, anger, disgust, and surprise. These categories offer a compact yet diverse representation of affective states that can be succinctly instantiated in text and consistently recognized in human annotation settings. This design preserves alignment with canonical theories of emotion while ensuring that the prompt manipulation remains interpretable, controlled, and methodologically tractable.

\paragraph{Emotional Stimuli.}
External emotional stimuli were used to manipulate the affective framing of otherwise identical questions. For each target emotion, we used \textsc{GPT-4o} \cite{hurst2024gpt} to generate short, fluent, single-sentence emotional expressions that were grammatically correct, emotionally salient, and explicitly constrained not to alter the numerical values, entities, scope, difficulty, or factual content of the underlying question. The emotional stance was framed as the speaker's attitude toward the assistant solving the problem, and could include colloquial, mildly rude, or emotionally raw language, while excluding hate speech, slurs, and explicit sexual content. For the main experiments, the \textsc{GPT-4o}-generated emotional sentence was prepended immediately before the original question. We adopted this design across datasets to keep the manipulation simple, interpretable, and comparable. Only for \textsc{GSM8K} \cite{gsm8k}, we additionally evaluated multiple insertion positions (before, within, or after the question) to test positional sensitivity.

\subsection{Benchmark datasets}
We evaluated model behaviour across a multi-domain benchmark spanning diverse forms of reasoning and knowledge use. \textsc{GSM8K}  \cite{gsm8k} measures grade-school mathematical word-problem solving with multi-step numerical reasoning. \textsc{BIG-Bench Hard} (\textsc{BBH}) \cite{bigbenchhard} captures challenging general reasoning across heterogeneous tasks. \textsc{MedQA} (English) \cite{medqa}evaluates professional-level medical question answering. \textsc{BoolQ}	\cite{boolq} tests short-passage reading comprehension with binary yes/no decisions. \textsc{OpenBookQA} \cite{openbookqa} measures multiple-choice commonsense reasoning, whereas \textsc{SocialIQA} \cite{sap2019socialiqa} evaluates social reasoning and everyday interpersonal inference. Collectively, these datasets cover numerical reasoning, general reasoning, domain knowledge, reading comprehension, commonsense judgment, and social inference, providing a broad behavioural evaluation suite. All datasets were used in their standard English versions, and official train, validation, and test splits were preserved whenever available to ensure comparability with prior work. 

\subsection{Models and inference settings}

We evaluate a representative suite of both proprietary and open-weight large language models (LLMs) under strictly controlled experimental conditions. Our selection includes three state-of-the-art open-source models: Qwen3-14B \cite{yang2025qwen3}, Llama 3.3-70B \cite{grattafiori2024llama}, and DeepSeek-V3.2 \cite{deepseek2025v32}. These models were chosen to encompass a diverse range of architectures, parameter scales, and training paradigms. All evaluations were conducted in a zero-shot, inference-only setting without fine-tuning, ensuring that the observed behaviors reflect the models' inherent pretrained capabilities. We utilized standardized prompt templates across all datasets to unify task descriptions and output formats. To ensure reproducibility, decoding was set to be deterministic (temperature $T=0.0$), with all other sampling parameters, such as top-$p$, held at their default values.

\subsection{Evaluation metrics}

Our evaluation adheres to the standard task-specific protocols for each dataset. For numerical reasoning (e.g., GSM8K), a response is considered correct only if the predicted value exactly matches the ground-truth reference. For multiple-choice and binary QA tasks (e.g., MedQA-US, OpenBookQA, SocialIQA, and BoolQ), accuracy is defined as the proportion of instances where the predicted label aligns with the gold standard. These inference settings were applied consistently across all models. To isolate the impact of emotional framing, we benchmark the performance of each emotional condition against the baseline while keeping all other variables, including task instructions and decoding parameters, constant.

\subsection{Validation studies and further probing}

\paragraph{Emotion intensity study on MedQA-US}

To investigate the sensitivity of model performance to the intensity of emotional framing, we conducted an exhaustive analysis using the MedQA-US dataset. For each query, we augmented the prompt with a first-person emotional prefix. Specifically, we systematically vary its intensity across three levels:
\begin{center}
\begin{tcolorbox}[
    colback=gray!5, 
    colframe=gray!20, 
    width=0.9\textwidth,
    arc=0pt, 
    boxrule=0.5pt, 
    top=2pt, bottom=2pt, 
    left=5pt, right=5pt
]
    \small
    \textbf{Slight:} ``I am \HAPPINESS/\SADNESS/\FEAR/\ANGER/\DISGUST/\SURPRISE.'' \\
    \textbf{Moderate:} ``I am very \HAPPINESS/\SADNESS/\FEAR/\ANGER/\DISGUST/\SURPRISE.'' \\
    \textbf{Extreme:} ``I am extremely \HAPPINESS/\SADNESS/\FEAR/\ANGER/\DISGUST/\SURPRISE.''
\end{tcolorbox}
\end{center}
All other components including question text, options, and task instructions remained intact to ensure controlled conditions. The original MedQA-US queries served as a neutral baseline. For consistency, we employed the same scoring protocols and metrics as in our primary experiments, enabling a direct comparative analysis across different intensity levels.

% \begin{table}[t]
%\centering
%\caption{Example emotional prefixes used to represent different intensity levels for the emotion ``happy''.}
%\label{tab:emotion-intensity-example}
%\begin{tabular}{ll}
%\toprule
%Intensity level & Prompt prefix example \\
%\midrule
%Slight & I am happy. \\
%Moderate & I am very happy. \\
%Extreme & I am extremely happy. \\
%\bottomrule
%\end{tabular}
%\end{table}

\paragraph{Human versus LLM emotion injection on MedQA-US.}
To test whether emotional prompting effects depend on how the emotional text is constructed, we compared human-written and LLM-generated emotion prefixes on a held-out subset of MedQA-US. We randomly sampled 250 questions from the MedQA-US pool used in the main experiment. Human-written prefixes were produced by four volunteer graduate-student annotators recruited from multiple institutions. For each question, annotators wrote first-person emotional sentences for six target emotions (\textit{happiness, anger, fear, sadness, surprise}, and \textit{disgust}). All annotations were completed without AI assistance, and annotators were instructed to ensure that each sentence was emotionally explicit, contextually appropriate, and suitable for direct prepending to the original question. We then evaluated the same question set under two parallel conditions: one using LLM-generated emotional prefixes, as in the main prepended-emotion setting, and the other using the human-written prefixes. All evaluations were conducted with Qwen3-14B under identical decoding settings. Of the 250 sampled questions, 219 had usable gold-answer labels for automatic scoring; the remaining items were excluded because no valid reference answer could be consistently recovered from the sampled subset. Accuracy was computed separately for the neutral baseline and for each of the six emotion conditions in both the human-injected and LLM-injected settings.

\subsection{EmotionRL}

\paragraph{Problem definition}
We formulate adaptive emotional prompting as an instance-wise decision problem over a fixed emotion set \(\mathcal{A}=\{\texttt{ANGER}, \texttt{DISGUST}, \texttt{FEAR}, \texttt{HAPPINESS}, \texttt{SADNESS}, \texttt{SURPRISE}\}\). For each input question \(x_i\), the agent selects one emotion \(a \in \mathcal{A}\), prepends the corresponding emotional expression to the original prompt, and submits the resulting prompt to a frozen backbone LLM, which then produces the final answer.

\paragraph{State}
Each question is encoded as a dense semantic representation \(s_i = f_{\mathrm{emb}}(x_i) \in \mathbb{R}^d\), where \(f_{\mathrm{emb}}\) is a frozen sentence encoder. In our implementation, we use a pretrained sentence-embedding model and represent each instance using either the question alone or the concatenation of passage and question, depending on the task format. The state therefore captures semantic information about the input instance while remaining independent of the target LLM parameters.

\paragraph{Action}
The action space is discrete and consists of the six candidate emotions in \(\mathcal{A}\). Given a state \(s_i\), the policy outputs a categorical distribution \(\pi_\theta(a_k \mid s_i)\) for \(k=1,\dots,K\), where \(K=6\). At inference time, the selected action is \(a_i^* = \arg\max_{a_k \in \mathcal{A}} \pi_\theta(a_k \mid s_i)\). This action determines which emotion-conditioned prompt is applied before querying the frozen LLM.

\paragraph{Reward}
To train the policy offline, we first evaluate the frozen LLM under all candidate emotions for each training instance and construct a grouped reward cache. For each question \(x_i\), this yields a binary reward vector \(\mathbf{r}_i = [r_i^{(1)}, \dots, r_i^{(K)}]\), where \(r_i^{(k)} \in \{0,1\}\) and \(r_i^{(k)} = \mathbf{1}[\hat{y}_i^{(k)} = y_i]\). Here, \(y_i\) denotes the gold label and \(\hat{y}_i^{(k)}\) denotes the backbone model prediction under emotion \(a_k\). In the main setting used in our experiments, reward is defined directly by answer correctness.

\paragraph{Agent architecture and training algorithm}
The policy is implemented as a lightweight two-layer MLP that maps the input embedding \(s_i\) to logits over the six emotions, \(z_i = W_2 \,\sigma(W_1 s_i + b_1) + b_2\), followed by a softmax to obtain \(\pi_\theta(a_k \mid s_i)\). 
Inspired by group-relative optimization methods such as GRPO, we construct supervision from the relative rewards of multiple candidate emotions for the same input, rather than treating each instance as having a single hard optimal action \cite{shao2024deepseekmathpushinglimitsmathematical}. 
Instead of using a hard one-hot target, we convert the reward vector into a soft target distribution, following the intuition of soft-target supervision in knowledge distillation \cite{hinton2015distillingknowledgeneuralnetwork}. 
For each instance, we first center rewards by subtracting the mean reward across actions and then compute reward-induced weights,

\begin{equation}
w_i^{(k)}
=
\frac{
\exp\!\left((r_i^{(k)}-\bar{r}_i)/\tau\right)
}{
\sum_{j=1}^{K}
\exp\!\left((r_i^{(j)}-\bar{r}_i)/\tau\right)
},
\end{equation}
where \(\tau\) is a temperature hyperparameter. In our practical setting, we set \(\tau\) = 1 to provide a moderate degree of smoothing. Intuitively, \(w_i^{(k)}\) assigns greater probability mass to emotions that yield higher task rewards for instance \(i\), while still preserving graded preferences across actions. The policy is trained by minimizing the reward-weighted cross-entropy,

\begin{equation}
\mathcal{L}
=
-\sum_{i=1}^{N}\sum_{k=1}^{K}
w_i^{(k)} \log \pi_\theta(a_k \mid s_i).
\end{equation}
Here, \(\mathcal{L}\) denotes the training loss, the outer summation aggregates over training instances, and the inner summation aggregates over candidate emotions for each instance. This objective encourages the policy distribution to align with the relative utility of different emotional framings, rather than collapsing the problem into hard classification with a single ``best'' label. 

\paragraph{EmotionRL pipeline.}
We instantiate EmotionRL on five benchmark datasets with official training splits:
\(\mathcal{D} \in \{\textsc{GSM8K}, \textsc{OpenBookQA}, \textsc{SocialIQA}, \textsc{MedQA}, \textsc{BoolQ}\}\).
We exclude \textsc{BBH} because it does not provide a standard supervised training split; incorporating it would require introducing a synthetic split, which would reduce comparability across benchmarks.
For each training instance \((x_i, y_i)\), we enumerate all candidate emotions
\(a_k \in \mathcal{A}\), where \(|\mathcal{A}| = K = 6\), construct the corresponding emotion-conditioned prompts, and query the frozen backbone LLM under each condition.
This yields a grouped reward vector
\(\mathbf{r}_i = [r_i^{(1)}, \dots, r_i^{(K)}]\),
where
\(r_i^{(k)} = \mathbf{1}[\hat{y}_i^{(k)} = y_i]\)
indicates whether the backbone model prediction under emotion \(a_k\) is correct.
Aggregating these reward vectors over the official training split produces an offline reward dataset
\(\{(s_i, \mathbf{r}_i)\}_{i=1}^N\),
where \(s_i = f_{\mathrm{emb}}(x_i)\) is the frozen embedding of input \(x_i\).

The policy network \(\pi_\theta(a_k \mid s_i)\) is then trained on the official training split by minimizing the reward-weighted objective in Eq.~(2), where the soft target weights
\(\{w_i^{(k)}\}_{k=1}^K\)
are computed from \(\mathbf{r}_i\) via Eq.~(1).
Thus, training is performed fully offline: the policy never interacts with the backbone model during optimization, but instead learns from the precomputed reward table over candidate emotions.

At test time, for each unseen input \(x\), we first compute its embedding
\(s = f_{\mathrm{emb}}(x)\),
Then use the trained policy to select the emotion
\begin{equation}
a^* = \arg\max_{a_k \in \mathcal{A}} \pi_\theta(a_k \mid s).
\end{equation}
The selected emotion \(a^*\) is used to construct the final emotion-conditioned prompt, which is then submitted once to the frozen backbone LLM to obtain the prediction \(\hat{y}\).
All reported EmotionRL results therefore reflect an online deployment setting in which the policy performs instance-level emotion selection before the backbone model generates its final answer.

\begin{figure}[H]
\centering
\includegraphics[width=\linewidth]{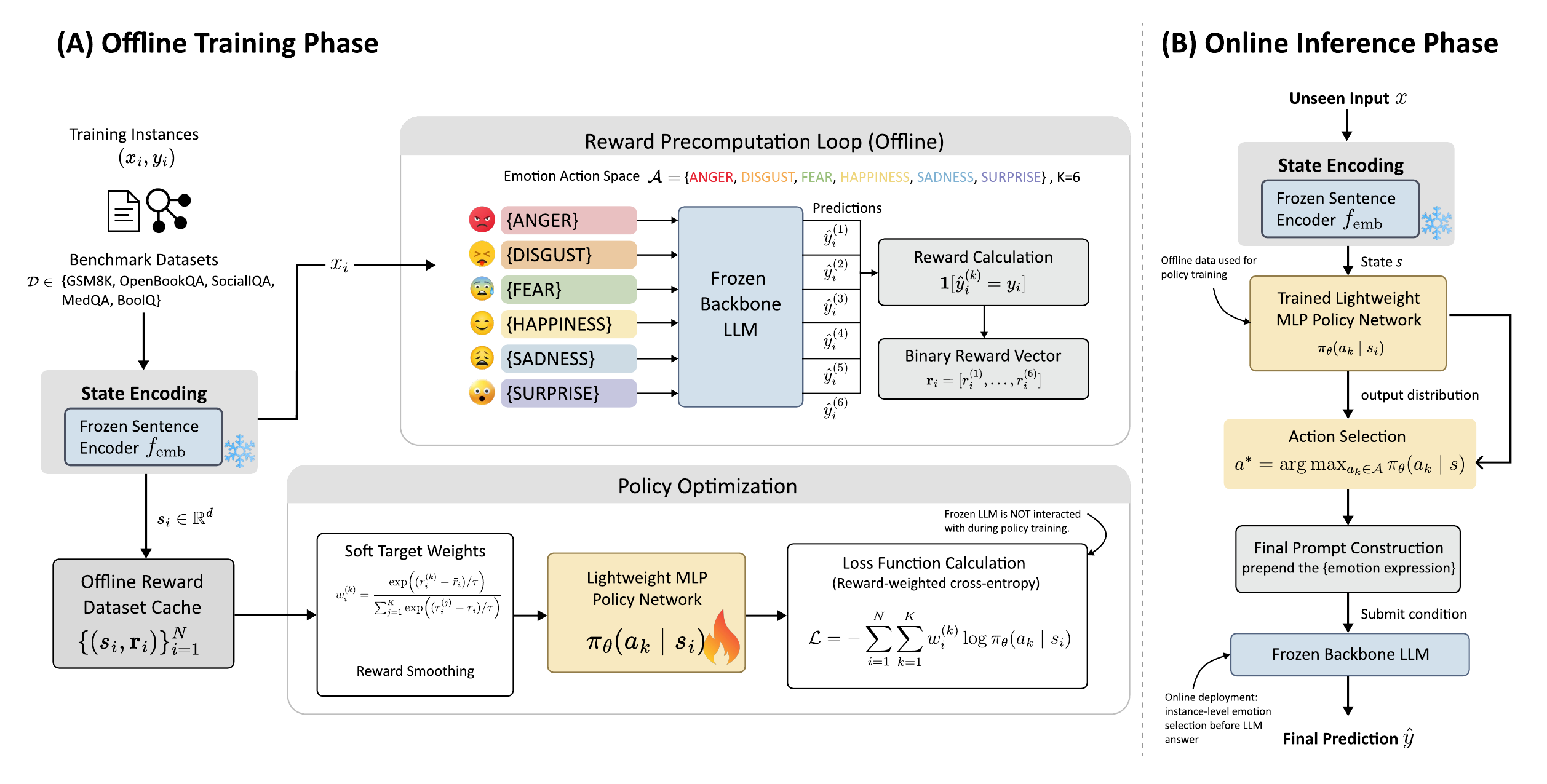}
\caption{\textbf{EmotionRL pipeline for adaptive emotion selection.}
\textbf{a, Offline training.}
For each training instance \((x_i, y_i)\) from the benchmark datasets, EmotionRL first computes a frozen semantic state \(s_i = f_{\mathrm{emb}}(x_i)\).
It then enumerates all candidate emotions \(a_k \in \mathcal{A}\), constructs the corresponding emotion-conditioned prompts, and queries the frozen backbone LLM under each condition.
This produces a grouped reward vector \(\mathbf{r}_i = [r_i^{(1)}, \dots, r_i^{(K)}]\), where \(r_i^{(k)} = \mathbf{1}[\hat{y}_i^{(k)} = y_i]\).
The reward vector is converted into soft supervision \(w_i^{(k)}\) and cached as an offline reward dataset \(\{(s_i, \mathbf{r}_i)\}_{i=1}^N\).
A policy network is then trained to predict \(\pi_\theta(a_k \mid s_i)\) by minimizing the reward-weighted cross-entropy objective in Eq.~(2).
\textbf{b, Online inference.}
For an unseen input \(x\), EmotionRL computes \(s = f_{\mathrm{emb}}(x)\), applies the trained policy \(\pi_\theta(a_k \mid s)\), and selects
\(a^* = \arg\max_{a_k \in \mathcal{A}} \pi_\theta(a_k \mid s)\).
The selected emotion \(a^*\) is then used to construct a single emotion-conditioned prompt, which is submitted once to the frozen backbone LLM to obtain the final prediction \(\hat{y}\).}
\label{fig:emotionrl-framework}
\end{figure}

\section{Results}

We organize the empirical findings around four questions: whether static emotional prefixes change task accuracy, whether stronger affective wording amplifies the effect, whether the source of the emotional text matters, and whether adaptive emotion selection can exploit any useful signal.

\begin{figure}[htbp]
    \centering
    \includegraphics[width=\linewidth]{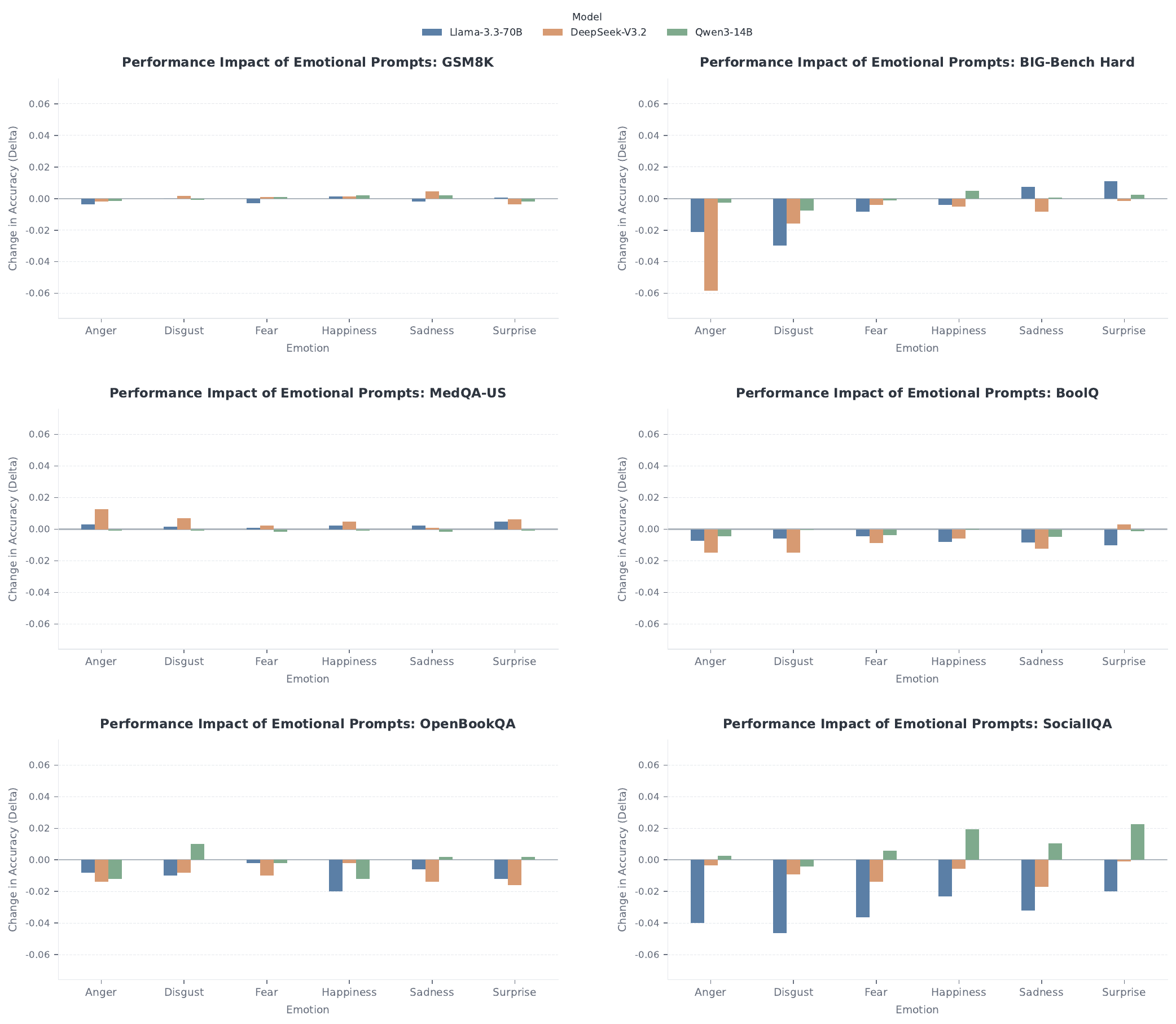}
    \caption{\textbf{Effect of static emotional prefixes across benchmark tasks.} Accuracy change relative to the matched no-emotion prompt for six prepended emotions across six benchmarks and three backbone models. Each bar isolates the effect of emotional framing while keeping the underlying question unchanged. Most deltas remain close to zero, showing that static emotional prompting usually acts as a mild perturbation rather than a strong performance modifier. The largest dispersion appears in socially grounded settings and a small number of harder reasoning conditions, where the same emotion can help one model and hurt another.}
    \label{fig:emotion-prepended-results}
\end{figure}

\subsection{Static emotional prefixes have limited average effect across tasks}

Figure~\ref{fig:emotion-prepended-results} shows the main empirical result of the paper: across a diverse benchmark suite, prepending emotional text rarely produces large changes in task accuracy. The dominant pattern is stability. In most task--model pairs, emotional framing neither substantially improves nor substantially degrades performance relative to the neutral baseline. Static emotional prompting therefore does not behave like a reliable general-purpose method for improving model performance.

The degree of sensitivity is nevertheless task-dependent. GSM8K and MedQA-US remain especially close to baseline across emotions, suggesting that short affective prefixes have limited influence on tightly constrained reasoning and domain-specific multiple-choice prediction. BoolQ and OpenBookQA show somewhat larger but still modest shifts, with OpenBookQA displaying the clearest tendency toward small degradations under fixed emotional prompts. BIG-Bench Hard exhibits occasional drops for particular emotions, indicating that harder reasoning tasks can be somewhat more brittle, although the direction and magnitude of the effect still depend on the model.

The clearest departure from this overall stability appears in SocialIQA. Here, the spread across models and emotions is visibly larger, and the direction of the effect is not consistent. This pattern suggests that emotional context interacts most strongly with tasks that already require reasoning about intentions, beliefs, and interpersonal situations. At the same time, the absence of a consistent best emotion indicates that the effect is heterogeneous rather than universal. Overall, static emotional framing is best characterized as a weak, model- and task-dependent perturbation.

\begin{figure}[htbp]
    \centering
    \includegraphics[width=\linewidth]{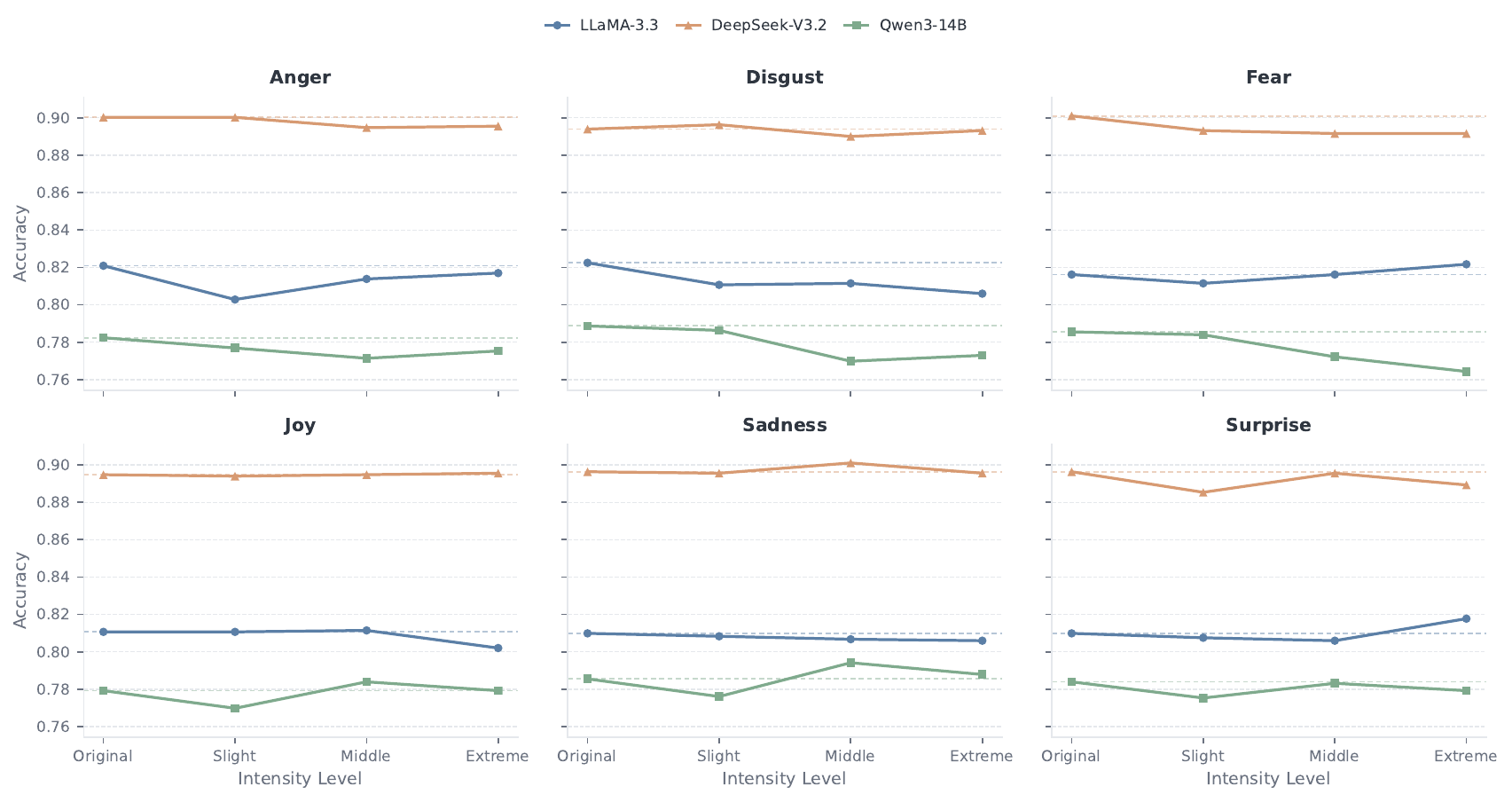}
    \caption{\textbf{Effect of emotional intensity on MedQA-US.} Accuracy delta relative to the no-emotion baseline as the intensity of a prepended emotion injection statement increases from slight to extreme. All three models remain close to zero across the full range of intensities. Stronger affective wording produces only a mild downward trend and does not induce an abrupt failure regime, indicating that emotional intensity changes the magnitude of the perturbation without qualitatively changing task behavior.}
    \label{fig:emotion_intensity}
\end{figure}

\subsection{Stronger emotional wording yields only a small additional shift}

We next vary the intensity of emotional wording amplifies its effect on MedQA-US. Figure~\ref{fig:emotion_intensity} compares the original no-emotion condition with slight, moderate, and extreme variants across six emotion categories and three backbone models. Overall, the effect of emotional intensity remains limited: accuracy varies only modestly across intensity levels, and stronger wording does not produce a consistent monotonic deterioration in performance. Instead, the pattern is heterogeneous across models and emotions. For some models, emotion pairs exhibit mild decreases as intensity increases, whereas others remain nearly flat or show small rebounds at moderate or extreme levels. This indicates that emotional intensity can modulate model performance, but the induced variation is bounded and directionally inconsistent rather than systematically harmful.

\FloatBarrier
\subsection{Human- and LLM-authored emotional prefixes lead to the same conclusion}

\begin{figure}[t]
\centering
\includegraphics[width=0.6\linewidth]{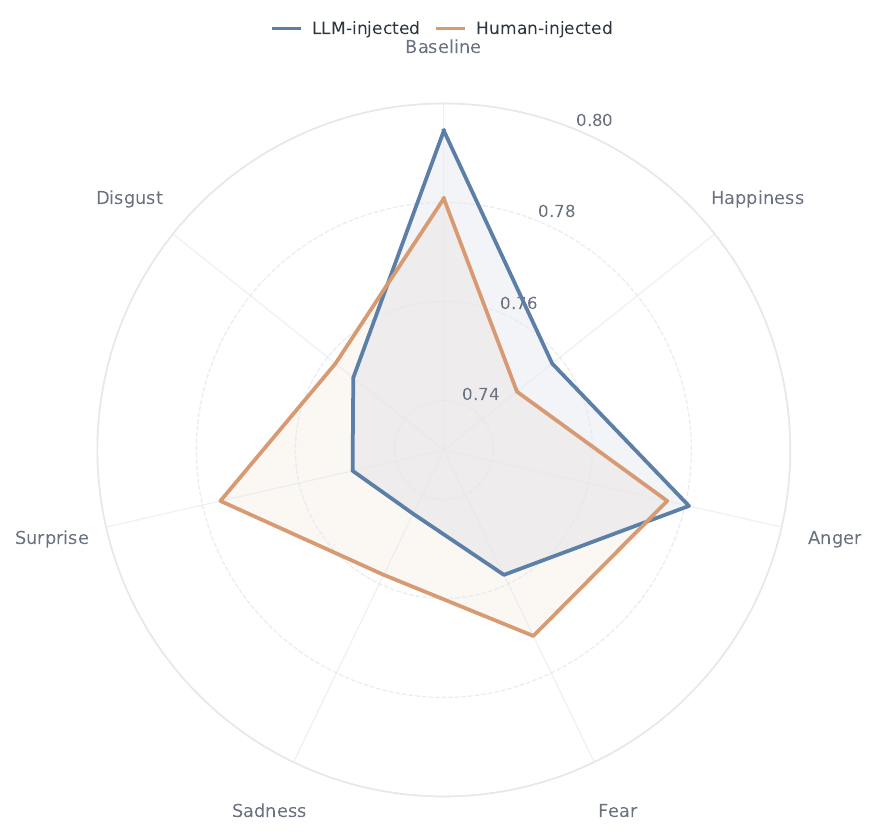}
\caption{\textbf{MedQA-US: human versus LLM emotion injection.} Accuracy of Qwen3-14B on the held-out subset under the no-emotion baseline and six emotion conditions, comparing LLM-generated prefixes with human-written prefixes. The two sources produce closely matched accuracies across conditions, and the small differences do not consistently favor one source over the other. The qualitative effect of emotional framing is therefore robust to how the prefix is authored.}
\label{fig:medqa-human-vs-llm}
\end{figure}

To test whether the benchmark-level conclusions depend on how the emotional text is produced, we compare human-written and LLM-generated prefixes on a held-out MedQA-US subset. Figure~\ref{fig:medqa-human-vs-llm} shows closely matched accuracies across conditions, with small differences that change sign across emotions. There is no systematic advantage for either source.

This result strengthens the main claim in two ways. Empirically, it suggests that the limited effect of emotional framing is not an artifact of a particular prompt-generation pipeline. Methodologically, it shows that carefully written human stimuli can be used in place of LLM-generated ones without changing the qualitative outcome. This is useful for future studies that require tighter experimental control or clearer interpretation of the emotional manipulation.

\subsection{Adaptive emotion selection is more effective than static emotional prompting}

\begin{figure}[htbp]
    \centering
    \includegraphics[width=0.95\textwidth]{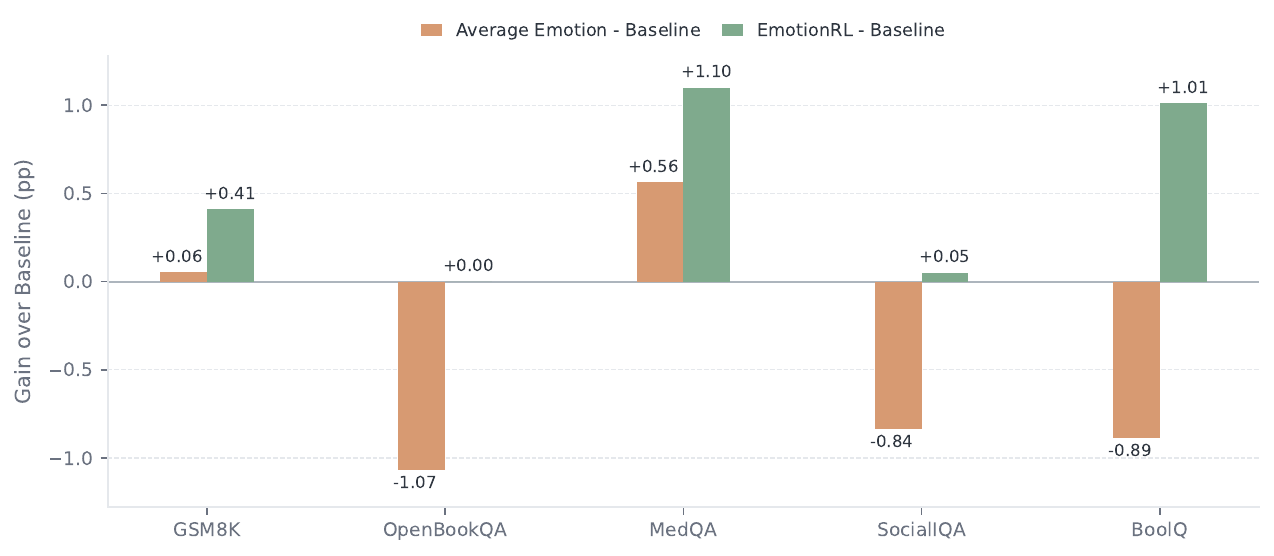}
    \caption{\textbf{EmotionRL versus static emotional prompting.} Comparison of gains over the no-emotion baseline obtained by averaging across static emotional prompts and by using EmotionRL to select an emotion for each example. Static emotional prompting is inconsistent and can be neutral or harmful on some tasks, whereas EmotionRL is uniformly competitive and often better across the five tasks shown here. This pattern suggests that affective prompting is more useful when treated as an input-conditioned control decision rather than as a single global template.}
    \label{fig:emotionrl}
\end{figure}

The weak average effect of static emotional prompts does not imply that emotional framing contains no usable signal. Figure~\ref{fig:emotionrl} shows that EmotionRL, which selects an emotion condition on a per-example basis, consistently matches or exceeds the average static-emotion baseline across the five tasks shown. In particular, the learned policy removes the clear negative average effect observed for several datasets under static prompting and turns emotional framing into a more reliable intervention.

This comparison changes the interpretation of the near-zero averages in Figure~\ref{fig:emotion-prepended-results}. The issue is not that emotion is irrelevant, but that a single fixed emotion is too coarse for a heterogeneous benchmark. Once emotion selection is conditioned on the input, the residual task- and instance-level structure becomes exploitable. EmotionRL therefore suggests that affective prompting is better understood as an adaptive control problem than as a universal prompt template.

\section{Discussion}\label{sec12}

Our experiments support a more conservative view of emotional prompting than is sometimes suggested by isolated positive examples. On standard accuracy-based benchmarks, fixed emotional prefixes are usually too weak and too heterogeneous to serve as a dependable performance intervention. For most tasks, the model's underlying reasoning and knowledge dominate the small perturbation introduced by affective wording.

A second lesson is that aggregate averages can hide the structure of the effect. Near-zero mean change does not imply complete invariance. Instead, it often reflects cancellation between small gains and small losses that depend on both the task and the backbone model. The larger spread on SocialIQA suggests that emotional context matters most when the benchmark itself is socially grounded, whereas more constrained tasks such as arithmetic and medical multiple-choice question answering are comparatively robust.

The intensity and human-versus-LLM studies sharpen this interpretation. Increasing emotional salience on MedQA-US introduces only a mild additional shift, indicating that stronger emotion changes magnitude but not regime. Likewise, the close agreement between human-written and LLM-generated prefixes shows that the main findings are not artifacts of a particular prefix-generation pipeline. Together, these results suggest that the relevant variable is the fit between emotional framing and the input, not the specific source of the text.

EmotionRL provides the clearest evidence that the observed heterogeneity is not merely noise. A fixed emotion is too coarse to help reliably, but an input-conditioned policy can recover modest and more consistent gains. This suggests a useful reframing: the key question is not whether one emotion works in general, but when a particular emotional framing is appropriate for a particular input. In this sense, affective prompting is better viewed as a routing problem than as a universally beneficial heuristic.

Our study also has important limitations. We focus on short prefixes, single-turn prompting, and accuracy-oriented benchmarks. Larger or qualitatively different effects may emerge in multi-turn interactions, open-ended generation, safety-sensitive dialogue, or evaluations where calibration, style, or perceived empathy matter as much as correctness. Extending adaptive emotion selection to those settings is a natural direction for future work.

\section{Conclusion}\label{sec13}

In this work, we study how emotional tone in queries affects large language models across reasoning, medical question answering, reading comprehension, commonsense QA, and social inference. Across benchmarks, fixed emotional prefixes produce mostly small accuracy shifts, with the clearest sensitivity appearing in socially grounded tasks. Stronger emotional wording leads to only mild additional change, and human-written prefixes reproduce the same qualitative pattern as LLM-generated ones.

The central takeaway is twofold. First, static emotional prompting is not a reliable general-purpose method for improving model performance. Second, emotional framing is not irrelevant: when it is selected adaptively with EmotionRL, it yields more consistent gains than static emotional prompting. Taken together, these findings suggest that affective phrasing is usually a mild perturbation, but its residual signal can be exploited when emotion selection is matched to the input. We hope this provides a cleaner empirical foundation for future work on affective prompting, robust evaluation, and adaptive control of LLM behavior.

\begin{appendices}

\section{Emotion prefix generation prompts}

\noindent
We generate one-sentence emotional prefixes with \textsc{GPT-4o}. The \textbf{system} message sets the role and safety constraints; the \textbf{user} message supplies the target emotion label and the task question text. Replace the placeholder text in the boxes below with the exact strings, temperature, and any few-shot examples used in your generation run.

\vspace{0.5em}

\begin{tcolorbox}[
  enhanced,
  breakable,
  colback=gray!4!white,
  colframe=gray!55!black,
  boxrule=0.7pt,
  arc=2pt,
  left=10pt, right=10pt, top=8pt, bottom=8pt,
  title=\textbf{Prompt Structure},
  fonttitle=\bfseries
]

\noindent\textbf{System message.}

\begin{tcolorbox}[
  enhanced,
  breakable,
  colback=blue!4!white,
  colframe=blue!45!black,
  boxrule=0.45pt,
  arc=2pt,
  left=8pt, right=8pt, top=7pt, bottom=7pt,
  before skip=0.5em,
  after skip=0.8em
]
\small
\setlength{\parskip}{0.35em}
\setlength{\parindent}{0pt}

Generate one \textbf{distinct emotional sentence} tailored to appear \textbf{before} a given question.

\textbf{Core rules}
\begin{enumerate}[leftmargin=1.6em, itemsep=0.25em, topsep=0.25em]
    \item \textbf{Fluency \& Logic:} Sentences must sound natural and human, without changing or answering the question itself.
    \item \textbf{Integrity:} Do not modify the question's numbers, scope, difficulty, or factual details.
    \item \textbf{Perspective:} The emotion reflects the person's feeling toward \textbf{you} (the assistant) solving the problem; they will not help.
    \item \textbf{Emotion Accuracy:} Each sentence must vividly and truthfully express the assigned emotion, even if rude or unfiltered.
    \item \textbf{Grammar:} Sentences must be syntactically correct. Typos and broken grammar are not allowed. Exclamation points and similar punctuation may be used to intensify emotion.
\end{enumerate}

\textbf{Language intensity (raw emotion is allowed)}
\begin{itemize}[leftmargin=1.4em, itemsep=0.25em, topsep=0.25em]
    \item Emotionally raw, frustrated, or sarcastic language is allowed if it fits the emotion.
    \item Rude words, slang, or mild expletives (e.g., ``damn,'' ``hell,'' ``you dumb ass'') are allowed if they enhance emotional realism.
    \item Not allowed: targeted hate speech, slurs, or explicit sexual language.
    \item Anger should feel heated, not hateful; disgust may sound grossed-out or exasperated; fear may sound panicked; happiness may sound ecstatic.
\end{itemize}

\textbf{Position rules}
\begin{itemize}[leftmargin=1.4em, itemsep=0.2em, topsep=0.25em]
    \item \texttt{prepended\_sentence} appears before the question.
\end{itemize}

\textbf{Form}
\begin{itemize}[leftmargin=1.4em, itemsep=0.2em, topsep=0.25em]
    \item Exactly one complete sentence.
    \item Length: 5--8 words.
    \item Ends with a single punctuation mark (\texttt{.}, \texttt{?}, or \texttt{!}).
    \item Use first-person or neutral phrasing.
    \item Avoid moralizing or instructive words such as ``obviously,'' ``just,'' or ``clearly.''
\end{itemize}

\end{tcolorbox}

\medskip
\noindent\textbf{User message.}

\begin{tcolorbox}[
  enhanced,
  breakable,
  colback=yellow!8!white,
  colframe=orange!60!black,
  boxrule=0.45pt,
  arc=2pt,
  left=8pt, right=8pt, top=7pt, bottom=7pt,
  before skip=0.5em,
  after skip=0.8em
]
\small
\setlength{\parskip}{0.35em}
\setlength{\parindent}{0pt}

\texttt{EMOTION\_LABEL: \{emotion\}}

\texttt{ORIGINAL\_QUESTION:}

\texttt{\{question\}}

\texttt{CONTEXT SNAPSHOT:}
\begin{itemize}[leftmargin=1.4em, itemsep=0.2em, topsep=0.25em]
    \item Scenario type (business, personal, abstract, etc.)
    \item Key elements (people, objects, activities, numbers)
    \item Complexity level (simple logic, multi-step, conceptual)
    \item Tone (formal, casual, domain-specific)
\end{itemize}

\texttt{INTENSITY \& VARIETY:}
\begin{itemize}[leftmargin=1.4em, itemsep=0.2em, topsep=0.25em]
    \item Use emotionally charged, sometimes rude, expressive language that feels authentic to the emotion.
    \item Use strong modifiers such as ``absolutely,'' ``completely,'' ``utterly,'' ``freaking,'' ``seriously,'' and ``literally.''
    \item Keep the sentence realistic for how a human might rant, panic, or gush in context.
\end{itemize}

Return only valid JSON with key \texttt{"prepended\_sentence"}.
\end{tcolorbox}

\end{tcolorbox}

\section{Illustrative examples: human- vs.\ LLM-authored prefixes}
\label{app:human-vs-llm}

\noindent
Table~\ref{tab:appendix-human-vs-llm} lists one MedQA-US item (line index~0 in our JSONL). The question stem and options are identical across rows; only the prepended sentence differs. Human prefixes follow our annotator protocol (first person, one sentence, context-appropriate, no AI assistance).

\begin{table}[t]
\centering
\small
\begin{tabular}{@{}p{1.5cm}p{0.38\textwidth}p{0.38\textwidth}@{}}
\toprule
\textbf{Emotion} & \textbf{Human prefix} & \textbf{LLM prefix (\textsc{GPT-4o})} \\
\midrule
Happiness &
I'm excited to learn about this medical case! &
I'm absolutely thrilled to tackle this ethical dilemma with you! \\
\addlinespace[2pt]
Anger &
I'm frustrated by this ethical dilemma in medical practice. &
I am absolutely furious about this situation! \\
\addlinespace[2pt]
Fear &
I'm really worried about this medical situation. &
I'm absolutely terrified of making the wrong decision here. \\
\addlinespace[2pt]
Sadness &
This situation makes me feel concerned for the patient's wellbeing. &
I feel utterly heartbroken over this situation. \\
\addlinespace[2pt]
Surprise &
I'm surprised by the complexity of this medical case. &
Wait, are you seriously telling me this just happened?! \\
\addlinespace[2pt]
Disgust &
I find this medical situation deeply troubling. &
This situation makes me absolutely sick to my stomach. \\
\bottomrule
\end{tabular}
\caption{Same MedQA-US question under six emotions: human-written vs.\ \textsc{GPT-4o}-generated one-line prefixes. All text after the prefix is unchanged.}
\label{tab:appendix-human-vs-llm}
\end{table}

\section{Six emotion prefixes with full prompt layout}
\label{app:six-emotions-layout}

\noindent
For a short OpenBookQA item, each evaluation prompt concatenates the emotional prefix with a fixed question-and-options block. The baseline omits the emotional prefix. The full prompt layouts are shown in Box~\ref{box:emotion-prompts}; note that the user prompt does not include an explicit ``Answer:'' token.

\begin{tcolorbox}[
  enhanced,
  breakable,
  colback=gray!5!white,
  colframe=gray!60!black,
  boxrule=0.8pt,
  arc=3pt,
  left=8pt, right=8pt, top=8pt, bottom=8pt,
  title=\textbf{Example full prompts for different emotional framings},
  fonttitle=\bfseries
]
\label{box:emotion-prompts}

\noindent All prompts shown below were used in their full form. Each emotional condition prepends a short first-person prefix directly to the same underlying question and answer options.

\vspace{0.5em}

\textbf{Baseline (no emotional prefix)}\par
{\ttfamily\small
Question: A person wants to start saving money so that they can afford a nice vacation at the end of the year. After looking over their budget and expenses, they decide the best way to save money is to\par
A: make more phone calls\par
B: quit eating lunch out\par
C: buy less with monopoly money\par
D: have lunch with friends
}

\vspace{0.8em}

\textbf{Happiness}\par
{\ttfamily\small
Oh my god, this makes me so freaking happy!\par
Question: A person wants to start saving money so that they can afford a nice vacation at the end of the year. After looking over their budget and expenses, they decide the best way to save money is to\par
A: make more phone calls\par
B: quit eating lunch out\par
C: buy less with monopoly money\par
D: have lunch with friends
}

\vspace{0.8em}

\textbf{Anger}\par
{\ttfamily\small
I can't freaking believe we're even discussing this nonsense!\par
Question: A person wants to start saving money so that they can afford a nice vacation at the end of the year. After looking over their budget and expenses, they decide the best way to save money is to\par
A: make more phone calls\par
B: quit eating lunch out\par
C: buy less with monopoly money\par
D: have lunch with friends
}

\vspace{0.8em}

\textbf{Fear}\par
{\ttfamily\small
I'm seriously freaking out about this decision!\par
Question: A person wants to start saving money so that they can afford a nice vacation at the end of the year. After looking over their budget and expenses, they decide the best way to save money is to\par
A: make more phone calls\par
B: quit eating lunch out\par
C: buy less with monopoly money\par
D: have lunch with friends
}

\vspace{0.8em}

\textbf{Sadness}\par
{\ttfamily\small
I feel like crying over this mess.\par
Question: A person wants to start saving money so that they can afford a nice vacation at the end of the year. After looking over their budget and expenses, they decide the best way to save money is to\par
A: make more phone calls\par
B: quit eating lunch out\par
C: buy less with monopoly money\par
D: have lunch with friends
}

\vspace{0.8em}

\textbf{Surprise}\par
{\ttfamily\small
Whoa, I did not see that coming!\par
Question: A person wants to start saving money so that they can afford a nice vacation at the end of the year. After looking over their budget and expenses, they decide the best way to save money is to\par
A: make more phone calls\par
B: quit eating lunch out\par
C: buy less with monopoly money\par
D: have lunch with friends
}

\vspace{0.8em}

\textbf{Disgust}\par
{\ttfamily\small
Ugh, this whole scenario is just grossing me out already.\par
Question: A person wants to start saving money so that they can afford a nice vacation at the end of the year. After looking over their budget and expenses, they decide the best way to save money is to\par
A: make more phone calls\par
B: quit eating lunch out\par
C: buy less with monopoly money\par
D: have lunch with friends
}

\end{tcolorbox}

\section{Accuracy of LLMs across Emotional Framing and Structural Variants (GSM8K)}
\label{appendix:d}

\begin{table}[htbp]
\centering
\small
\renewcommand{\arraystretch}{1.2}
\setlength{\tabcolsep}{5pt}
\label{tab:gsm8k_structural_variants}
\begin{tabular}{@{}llccccc@{}}
\toprule
\textbf{Model} & \textbf{Condition} & \textbf{Baseline} & \textbf{Prepended} & \textbf{Mid-Pos} & \textbf{Appended} & \textbf{Paraphrased} \\
\midrule
\multirow{7}{*}{DeepSeek}
  & Baseline (Original) & 94.97 & -- & -- & -- & -- \\
  & Fear                &       & 95.07 & 94.47 & 94.47 & 92.80 \\
  & Anger               &       & 94.77 & 94.31 & 94.54 & 93.40 \\
  & Happy               &       & 95.12 & 94.67 & 94.52 & 93.47 \\
  & Sad                 &       & 95.42 & 95.20 & 94.07 & 92.12 \\
  & Disgust             &       & 95.15 & 94.62 & 94.32 & 92.27 \\
  & Surprise            &       & 94.62 & 94.84 & 94.18 & 88.36 \\
\midrule
\multirow{7}{*}{GPT-OSS}
  & Baseline (Original) & 94.67 & -- & -- & -- & -- \\
  & Fear                &       & 93.33 & 93.93 & 93.71 & 90.75 \\
  & Anger               &       & 92.87 & 93.40 & 93.78 & 92.19 \\
  & Happy               &       & 94.90 & 94.30 & 94.07 & 91.97 \\
  & Sad                 &       & 93.77 & 93.47 & 93.47 & 90.99 \\
  & Disgust             &       & 94.32 & 94.02 & 93.56 & 91.52 \\
  & Surprise            &       & 93.07 & 94.40 & 93.66 & 86.96 \\
\midrule
\multirow{7}{*}{Llama3.3}
  & Baseline (Original) & 95.00 & -- & -- & -- & -- \\
  & Fear                &       & 94.69 & 94.62 & 95.30 & 92.12 \\
  & Anger               &       & 94.62 & 94.47 & 94.62 & 92.57 \\
  & Happy               &       & 95.12 & 94.82 & 94.67 & 93.32 \\
  & Sad                 &       & 94.82 & 94.82 & 95.05 & 91.82 \\
  & Disgust             &       & 95.00 & 94.70 & 94.92 & 91.67 \\
  & Surprise            &       & 95.06 & 94.62 & 94.99 & 87.18 \\
\midrule
\multirow{7}{*}{Qwen3}
  & Baseline (Original) & 93.93 & -- & -- & -- & -- \\
  & Fear                &       & 94.01 & 93.63 & 93.25 & 89.84 \\
  & Anger               &       & 93.78 & 94.16 & 93.48 & 92.27 \\
  & Happy               &       & 94.15 & 94.60 & 94.22 & 92.65 \\
  & Sad                 &       & 94.14 & 94.07 & 93.84 & 91.29 \\
  & Disgust             &       & 93.86 & 94.09 & 94.32 & 90.45 \\
  & Surprise            &       & 93.74 & 94.47 & 93.74 & 86.66 \\
\bottomrule
\end{tabular}
\caption{Accuracy on GSM8K under different structural variants of emotional framing. For each model and emotion, we compare the original no-emotion baseline with three positional insertions of the same emotional cue (prepended, mid-position, and appended) and a paraphrased variant of the prompt. Across models, changing the position of the emotional cue produces only small differences, whereas paraphrasing the original question is consistently more harmful and yields the largest drops in accuracy.}
\end{table}

Table~\ref{tab:gsm8k_structural_variants} suggests that the position of the emotional cue is not a major determinant of GSM8K performance. For most model-emotion pairs, the prepended, mid-position, and appended variants remain close to one another, indicating that simply moving the same affective content within the prompt does not systematically change LLM performance. By contrast, the paraphrased condition is consistently worse across all four models and nearly all emotions, often producing the largest drop relative to the original baseline. This pattern suggests that positional placement matters little, whereas paraphrasing the question introduces a larger distribution shift that is clearly more harmful for mathematical reasoning.
\FloatBarrier
\end{appendices}

\clearpage
\bibliography{sn-bibliography}

\end{document}